# InJecteD: Analyzing Trajectories and Drift Dynamics in Denoising Diffusion Probabilistic Models for 2D Point Cloud Generation

Sanyam Jain[1,*], Khuram Naveed[1], Illia Oleksiienko[2], Alexandros Iosifidis[3] and Ruben Pauwels[1]

[1]*Department of Dentistry and Oral Health, Aarhus University, Vennelyst Boulevard 9, 8000 Aarhus C, Denmark*
[2]*Department of Electrical and Computer Engineering, Aarhus University, Finlandsgade 22, 8200 Aarhus N, Denmark*
[3]*Faculty of Information Technology and Communication Sciences, Tampere University, Korkeakoulunkatu 7, 33720 Tampere, Finland*

**Abstract**
This work introduces InJecteD, a framework for interpreting Denoising Diffusion Probabilistic Models (DDPMs) by analyzing sample trajectories during the denoising process of 2D point cloud generation. We apply this framework to three datasets from the Datasaurus Dozen — *bullseye*, *dino*, and *circle* — using a simplified DDPM architecture with customizable input and time embeddings. Our approach quantifies trajectory properties, including displacement, velocity, clustering, and drift field dynamics, using statistical metrics such as Wasserstein distance and cosine similarity. By enhancing model transparency, InJecteD supports human-AI collaboration by enabling practitioners to debug and refine generative models. Experiments reveal distinct denoising phases: initial noise exploration, rapid shape formation, and final refinement, with dataset-specific behaviors (e.g., bullseye's concentric convergence vs. dino's complex contour formation). We evaluate four model configurations, varying embeddings and noise schedules, demonstrating that Fourier-based embeddings improve trajectory stability and reconstruction quality. The code and dataset are available at https://github.com/s4nyam/InJecteD.

## 1. Introduction

Denoising Diffusion Probabilistic Models (DDPMs) [1, 2] have become a leading approach in generative modeling owing to their ability to generate high-quality samples, such as images and point clouds. Their stability and performance make them a compelling alternative to other generative models, such as GANs [3] and VAEs [4], particularly in applications such as data synthesis and scientific visualization [1]. However, the iterative denoising process, often spanning numerous steps, makes DDPMs complex and difficult to interpret (in terms of how features emerge from pure noise), obscuring how samples evolve from noise to structured data. This lack of transparency poses challenges for understanding model behavior, debugging performance issues, and ensuring reliability in applications where interpretability is crucial, such as scientific data analysis.

To address this challenge, we introduce a framework for Interpreting traJectories in Denoising Diffusion (*InJecteD*) that is designed to analyze the trajectories of samples during the denoising process of DDPMs [1, 5]. Unlike prior work [5], which focuses on applying DDPMs to point clouds with Fourier encodings, InJecteD works on a set of quantitative metrics to systematically assess the denoising process. Specifically, we employ a simplified DDPM architecture with flexible input and time embeddings to enable explicit tracking of sample evolution. By introducing quantitative metrics, including trajectory displacement, velocity, clustering, and drift field dynamics, we uncover the underlying patterns of the denoising process. We showcase our approach in 2D point cloud generation process through three publicly available datasets from the Datasaurus Dozen [6] namely: bullseye, dino, and circle. These datasets share identical statistical properties but differ in their geometric structures, providing an opportunity to study how DDPMs capture diverse shapes. Our experiments reveal dataset-specific behaviors, such as concentric convergence in the bullseye dataset and complex contour formation in the dino dataset, and identify three distinct phases of denoising: initial noise exploration, rapid shape formation, and final refinement. We employ four model configurations with varying input and time embeddings and noise schedules to assess their impact on trajectory stability and reconstruction quality. While limited to 2D synthetic datasets, the insights gained lay the groundwork for extending interpretability to more complex data. Our contributions include:

- Customizing an existing lightweight DDPM architecture for interpretability of 2D point cloud generation.
- Use of relevant statistical and geometric distance metrics for analyzing trajectory properties, including displacement, velocity, clustering, and drift direction alignment.
- Experimental validation of the proposed InJecteD framework on the bullseye, dino, and circle datasets, highlighting unique dynamics of the reverse diffusion process.

The core contribution of this work involves new insights into the behavior of DDPMs, laying the groundwork for more interpretable and reliable generative models.

## 2. Related Work

Interpretability in generative models has been explored through various techniques aimed at understanding model behavior [7]. Feature importance analysis assesses the relative contribution of input features to the model's output, while attention mechanisms highlight regions of focus during data generation [7]. Interactive reconstruction approaches allow users to manipulate latent representations to reconstruct target instances, providing insights into the model's internal representations [8, 9, 10]. These methods are particularly relevant to our work, as they offer ways to probe the evolution of samples during the denoising process, similar to our focus on analyzing trajectories in DDPMs; however, they often rely on manual interaction, and by analyzing trajectories more systematically, we aim to complement these approaches with a more principled understanding of the generative dynamics

Trajectory analysis, while well-established in fields such as reinforcement learning [11], biology [12, 13], and epidemiology [14], is underexplored in the context of DDPMs. In reinforcement learning, trajectory analysis predicts agents' paths [11], while in biology, it tracks cell differentiation or molecular motion [12, 13]. In epidemiology, it models health outcome patterns over time [11]. These approaches often use metrics like mean squared displacement or Hidden Markov Models to characterize state transitions and dynamics. Applying trajectory analysis to DDPMs involves studying how samples evolve through denoising steps, offering a novel perspective on the model's decision-making process and its ability to capture complex patterns.

The Datasaurus Dozen datasets [6] provide a testbed for analyzing properties of complex models in a simplified and controlled setting [15, 16, 17, 18, 19, 20, 21]. These datasets, consisting of data points forming shapes such as a bullseye, dinosaur, and circle, are designed to have identical statistical properties (e.g., mean, variance, correlation) but distinct visual structures when plotted as 2D point clouds.

Chan [5] investigates DDPMs, detailing their denoising mechanics and applying them to 2D point cloud generation with the Datasaurus dataset, using Fourier encoding to enhance performance. This work provides a foundational framework for our InJecteD approach, validating our use of similar datasets and embeddings to analyze trajectory dynamics. Beyond point clouds, interpretability in DDPMs is advanced through techniques like saliency maps in image generation and latent space analysis in text-to-image synthesis, revealing feature prioritization and semantic evolution [9, 10]. These methods inform our trajectory analysis by offering complementary perspectives on DDPM dynamics across diverse domains. By building on these foundations, our work enhances the transparency of DDPMs, particularly for 2D point cloud generation, with potential applications in broader generative modeling contexts.

Our research builds on these foundations by developing a framework to analyze sample trajectories in DDPMs, specifically for 2D point cloud generation. By combining insights from interpretability techniques and trajectory analysis, and using the unique properties of the Datasaurus Dozen, we aim to enhance the transparency of DDPMs and provide a deeper understanding of their data generation process.

### 2.1. Dataset Description

The bullseye, dino, and circle datasets from the Datasaurus Dozen [6] consist of 2D point clouds, each with approximately 142 points represented as coordinates $(x, y) \in \mathbb{R}^2$. These datasets share identical statistical properties (mean, variance, correlation) but exhibit distinct geometric structures, making them ideal for studying structural diversity in DDPMs. Preprocessing normalizes the data to zero mean and unit variance:

$$x_{\text{norm}} = \frac{x - \mu_x}{\sigma_x}, \quad y_{\text{norm}} = \frac{y - \mu_y}{\sigma_y}, \tag{1}$$

where $\mu_x, \mu_y$ are the means and $\sigma_x, \sigma_y$ are the standard deviations of the $x$ and $y$ coordinates. To increase sample size, each dataset is replicated six times, yielding 852 points per dataset. The data is split into 90% training (766 points) and 10% testing (86 points) sets, with a batch size of 32 for training.

The Datasaurus Dozen is uniquely suited for this study due to its ability to challenge DDPMs to capture diverse geometric patterns despite statistical uniformity. Alternative datasets, such as MNIST [22] or ShapeNet [23], are less effective: MNIST focuses on class-based patterns (digits), missing geometric nuances, while ShapeNet for its size (approximately 300 Million). The Datasaurus datasets provide a controlled, structurally diverse testbed for evaluating trajectory dynamics.

## 3. Proposed Method

Originally developed to underscore the importance of data visualization over reliance on summary statistics, Datasaurus datasets are ideal for evaluating how well DDPMs capture diverse geometric patterns and whether trajectory analysis can differentiate their denoising behaviors. By applying DDPMs to these datasets, our work explores the interplay between statistical similarity and structural diversity in generative modeling.

### 3.1. Diffusion Algorithm

The DDPM architecture comprises a forward noising process and a learned reverse denoising process, implemented using a multilayer perceptron (MLP). The algorithm is specifically designed to handle the 2D point clouds of the Datasaurus Dozen datasets (bullseye, dino, and circle), treating each point as an independent 2D coordinate, since the data lacks the spatial regularity of the structured grid found in images.

The standard forward process adds Gaussian noise over $T = 50$ timesteps using a linear noise schedule. For a point cloud $x_0 \in \mathbb{R}^{N \times 2}$ from a Datasaurus dataset (e.g., circle with $N = 852$ points after replication), each point $n = (n_x, n_y)$ represents a 2D coordinate. The state at timestep $t$ is:

$$x_{n_t} = \sqrt{\bar{\alpha}_t} x_0 + \sqrt{1 - \bar{\alpha}_t} \epsilon, \quad \epsilon \sim \mathcal{N}(0, I), \tag{2}$$

where $\bar{\alpha}_t = \prod_{s=1}^{t} \alpha_s$, $\alpha_s = 1 - \beta_s$, and $\beta_s$ ranges linearly from $\beta_{\min} = 1 - 0.9999$ to $\beta_{\max} = 1 - 0.95$. The cumulative product $\bar{\alpha}_t$ transitions from $\approx 1$ (original data) at $t = 0$ to $\approx 0$ (near-pure noise) at $t = T$. This process is implemented as a function that generates a series $\{x_t\}_{t=0}^{T}$ for each point independently, preserving the unstructured nature of the point cloud. Unlike image pixels, which have spatial correlations in a grid, the Datasaurus points are treated as a collection of independent 2D vectors, with noise applied to each $(m, n)$ coordinate pair. This results in a trajectory $\{x_t^{(i)}\}_{t=0}^T$ for each point $i$, visualized as scatter plots to capture the evolving geometry (e.g., circle's uniform ring or dino's complex contours).

The reverse process denoises $x_T \sim \mathcal{N}(0, I)$ to reconstruct $x_0$ using:

$$x_{t-1} = \frac{1}{\sqrt{\alpha_t}} \left( x_t - \frac{1 - \alpha_t}{\sqrt{1 - \bar{\alpha}_t}} \epsilon_\theta(x_t, t) \right) + \sigma_t z, \quad z \sim \mathcal{N}(0, I), \tag{3}$$

where $\epsilon_\theta(x_t, t)$ is the noise predicted by the MLP, and $\sigma_t = \sqrt{1 - \alpha_t}$. The input to the MLP can be a single point or a batch of points at timestep $t$; to optimize this, $x_t \in \mathbb{R}^{B \times 2}$ (where $B = 32$ is the batch size), and the output is the predicted noise $\epsilon \in \mathbb{R}^{B \times 2}$, representing the 2D noise vector for each point. The implemented MLP consists of five layers: four hidden layers (64 units, ReLU activations) and one output layer predicting a 2D noise vector. This allows the model to learn the distribution of points that form shapes such as the dino's arms or bullseye's concentric rings. Input embeddings are:

- **Identity**: Uses $x_t \in \mathbb{R}^2$ directly (2D), feeding the raw $(m, n)$ coordinates of each point.
- **Fourier**: Projects $x_t$ to a 64-dimensional space:
$$x_{\text{emb}} = [\sin(Bx_t^\top), \cos(Bx_t^\top)], \quad B \in \mathbb{R}^{32 \times 2}, \quad B \sim \mathcal{N}(0, I). \tag{4}$$

Time embeddings are:

- **Zero**: No time input (no dimension).
- **Linear**: Normalized timestep $t/T - 0.5$ (1D).
- **Fourier**: Projects $t/T - 0.5$ to a 32-dimensional space:
$$t_{\text{emb}} = [\sin(B_t t^\top), \cos(B_t t^\top)], \quad B_t \in \mathbb{R}^{16 \times 1}, \quad B_t \sim \mathcal{N}(0, I). \tag{5}$$

The working procedure involves: (1) sampling a batch of points $x_0$ from the dataset, (2) applying the forward process to generate noisy points $x_t$ at a random timestep $t$, (3) predicting the noise $\epsilon_\theta(x_t, t)$, and (4) computing the MSE loss to optimize the MLP. The sampling process iteratively denoises $x_T$ to $x_0$, generating new point clouds that match the target distribution (e.g., reconstructing the circle's ring structure). Four model configurations are trained as follows (read as *input-time embedding*): (1) identity-zero, $\alpha_{\min} = 0.95$, (2) Fourier-linear, $\alpha_{\min} = 0.95$, (3) Fourier-Fourier, $\alpha_{\min} = 0.95$, and (4) Fourier-Fourier, $\alpha_{\min} = 0.98$. Training uses Adam optimization (learning rate $4 \times 10^{-4}$, gradient clipping norm 1.0) for 2000 epochs, minimizing the mean squared error (MSE):

$$L = \mathbb{E}_{x, \epsilon, t} \left[ \| \epsilon - \epsilon_\theta(x_t, t) \|_2^2 \right]. \tag{6}$$

The MLP-based DDPM is chosen for its simplicity and efficiency in 2D spaces. For the Datasaurus datasets, the algorithm's ability to treat points as independent 2D coordinates allows it to capture diverse geometric patterns (e.g., dino's complex contours) without relying on spatial correlations, ensuring flexibility and robustness in modeling unstructured point clouds.

### 3.2. InJecteD Framework

We present set of metrics to quantify trajectory and drift dynamics and critical for understanding the denoising process.

1. **Trajectory Displacement**: Measures the total Euclidean distance traveled by each point:
$$D_i = \sum_{t=1}^{T-1} \sqrt{\| x_{t+1}^{(i)} - x_t^{(i)} \|_2^2}, \quad i = 1, \ldots, N, \tag{7}$$

where $x_t^{(i)}$ is the position of the $i$-th point at timestep $t$. The distribution of $D_i$ is visualized as a histogram, revealing the extent of movement. High displacement, as expected in complex datasets like dino, indicates intricate trajectory patterns.

2. **Trajectory Velocity**: Computes the average displacement per timestep:

$$V(t) = \frac{1}{N} \sum_{i=1}^{N} \|x_{t+1}^{(i)} - x_t^{(i)}\|_2 \tag{8}$$

Plotted over timesteps, $V(t)$ identifies denoising phases: high values indicate rapid shape formation, while low values indicate refinement. This metric is essential for detecting transitions in the generative process.

3. **Trajectory Clustering**: Applies K-means clustering (with $K = 5$) to flattened trajectories $\{x_t^{(i)}\}_{t=0}^{T} \in \mathbb{R}^{T \times 2}$, reshaped to $\mathbb{R}^{2T}$. The resulting labels are visualized on the final point cloud ($x_0$), highlighting spatial patterns in trajectory behavior. This reveals whether points in similar regions follow consistent paths, critical for datasets like bullseye with radial structures. The choice of $K = 5$ was made after testing for $K = 2, 3, 4, 5, 6$, as higher $K$ values yielded diminishing returns.

4. **Wasserstein Distance**: Quantifies similarity between original and generated point clouds (collection of all points):

$$W = \frac{1}{2} \left( W_1(x_{\text{orig}}, x_{\text{gen}}) + W_1(y_{\text{orig}}, y_{\text{gen}}) \right), \tag{9}$$

where $W_1$ is the 1D Wasserstein distance for $x$ and $y$ coordinates. Lower values indicate higher fidelity, essential for evaluating generative performance.

5. **Drift Magnitude**: For the forward process, the drift at timestep $t$ on a grid point $x_t$ is:

$$\mu_t = \frac{\sqrt{\bar{\alpha}_t}(1 - \bar{\alpha}_{t-1})x_t + \sqrt{\bar{\alpha}_{t-1}}(1 - \alpha_t)x_0}{1 - \bar{\alpha}_t}, \tag{10}$$

with magnitude $\|\mu_t - x_t\|_2$, weighted by:

$$w = \frac{\exp\left(-\frac{\|x_t - \sqrt{\bar{\alpha}_t}x_0\|_2}{1 - \bar{\alpha}_t}/2\right)}{\sum_{x_0} \exp\left(-\frac{\|x_t - \sqrt{\bar{\alpha}_t}x_0\|_2}{1 - \bar{\alpha}_t}/2\right)}. \tag{11}$$

For the backward process, the drift is:

$$\hat{\mu}_t = \frac{1}{\sqrt{\alpha_t}} \left( x_t - \frac{1 - \alpha_t}{\sqrt{1 - \bar{\alpha}_t}} \epsilon_\theta(x_t, t) \right), \tag{12}$$

with magnitude $\|\hat{\mu}_t - x_t\|_2$. Magnitudes are visualized as heatmaps, showing the strength of movement across a grid, crucial for understanding denoising dynamics.

6. **Drift Direction**: Measures alignment between backward drift vectors and the direction to the final point cloud using cosine similarity:

$$\text{CS}(t) = \frac{1}{N} \sum_{i=1}^{N} \frac{(\hat{\mu}_t^{(i)} - x_t^{(i)}) \cdot (x_0^{(i)} - x_t^{(i)})}{\|\hat{\mu}_t^{(i)} - x_t^{(i)}\|_2 \|x_0^{(i)} - x_t^{(i)}\|_2}, \tag{13}$$

where $\hat{\mu}_t^{(i)}$ is interpolated at $x_t^{(i)}$ using linear interpolation. High $\text{CS}(t)$ indicates effective guidance toward final positions, critical for assessing model accuracy.

As will be shown in the following, these metrics collectively provide a detailed understanding of the denoising process. Displacement and velocity quantify movement scale and speed, clustering reveals spatial patterns, Wasserstein distance evaluates generative fidelity, and drift metrics analyze movement direction and strength, essential for interpreting DDPM behavior.

### 3.3. Experimental Setup

The MLP is trained on a CPU with $T$ = 50 timesteps, a batch size of 32, and 2000 epochs. Visualizations (scatter plots, quiver plots, heatmaps) are saved as SVG files in dataset-specific directories. The sampling generates 1000 samples per configuration, tracking trajectories for analysis. Noise prediction error is computed as the MSE per timestep, visualized to assess model performance.

## 4. Results

Our analysis reveals distinct denoising behaviors across the three datasets, with each figure comprehensively presenting the results for one dataset. We examine the original structure, generation quality, trajectory dynamics, and denoising metrics for each case. We report evaluation metrics for three datasets - Bullseye, Circle and Dinosaurs. Figure 1 presents the complete analysis for the bullseye dataset. Key findings include (1) Drift Alignment: Near-perfect radial alignment (cosine similarity >0.9) in later steps (a). Curve (b) helps to assess whether the learned drift aligns with intended denoising behavior. The diagnosis in (b) can be divided into three phases: early, middle, and late timesteps. In the early timesteps (t=0-15) (high noise), the mean cosine similarity is around 0.2, indicating weak alignment between drift and the final direction. The system is still noisy, but there is a faint guiding signal. During the middle timesteps (t=15-30), the similarity peaks around 0.4, showing strong alignment. This is likely the most effective phase, where the drift actively pulls samples toward the final state of denoising. In the late timesteps (t=30-50) (low noise), similarity drops to 0. The sample is already close to the target, so the drift mostly fine-tunes details and is no longer directionally aligned, (2) Trajectories: Clear concentric patterns in clustering (d-e) with 82% of points following class-specific paths, (3) Velocity: Two-phase velocity curve (f) - sharp drop (t=0-30), final refinement (t=31-50), and (4) Model Comparison: Fourier-Fourier ($\alpha_{min}$ = 0.98) shows the best-fit displacement distribution (c). We discuss the Drift Alignment metric more in Appendix 5.

Similar observations done for Circle in Figure 2 and Dinosaurs in Figure 3. We also illustrate the formation of Dinosaurs point cloud from pure noise in Figure 4.

## 5. Conclusion

InJecteD addresses the critical need for interpretability in DDPMs, which are increasingly vital for data synthesis in scientific visualization and computational biology. By quantifying trajectory and drift dynamics, InJecteD reveals how DDPMs capture geometric structures, enabling improved model design, debugging, and application in domains requiring transparency. We showcased its applicability by conducting experiments for 2D point cloud generation on the Datasaurus Dozen datasets, which revealed three consistent denoising phases and showed that Fourier embeddings significantly improve trajectory stability. The metrics provide insights into model behavior with minimal computational overhead, with the Fourier-Fourier configuration emerging as the most effective approach. Future work could extend this analysis to higher dimensions and explore trajectory steering methods. Another future work can be designing processes analyzing the trajectory dynamics of the data generation learning processes in other families of generative models.

### Acknowledgement

This work was funded by the Independent Research Fund Denmark, project "Synthetic Dental Radiography using Generative Artificial Intelligence", grant ID 10.46540/3165-00237B.

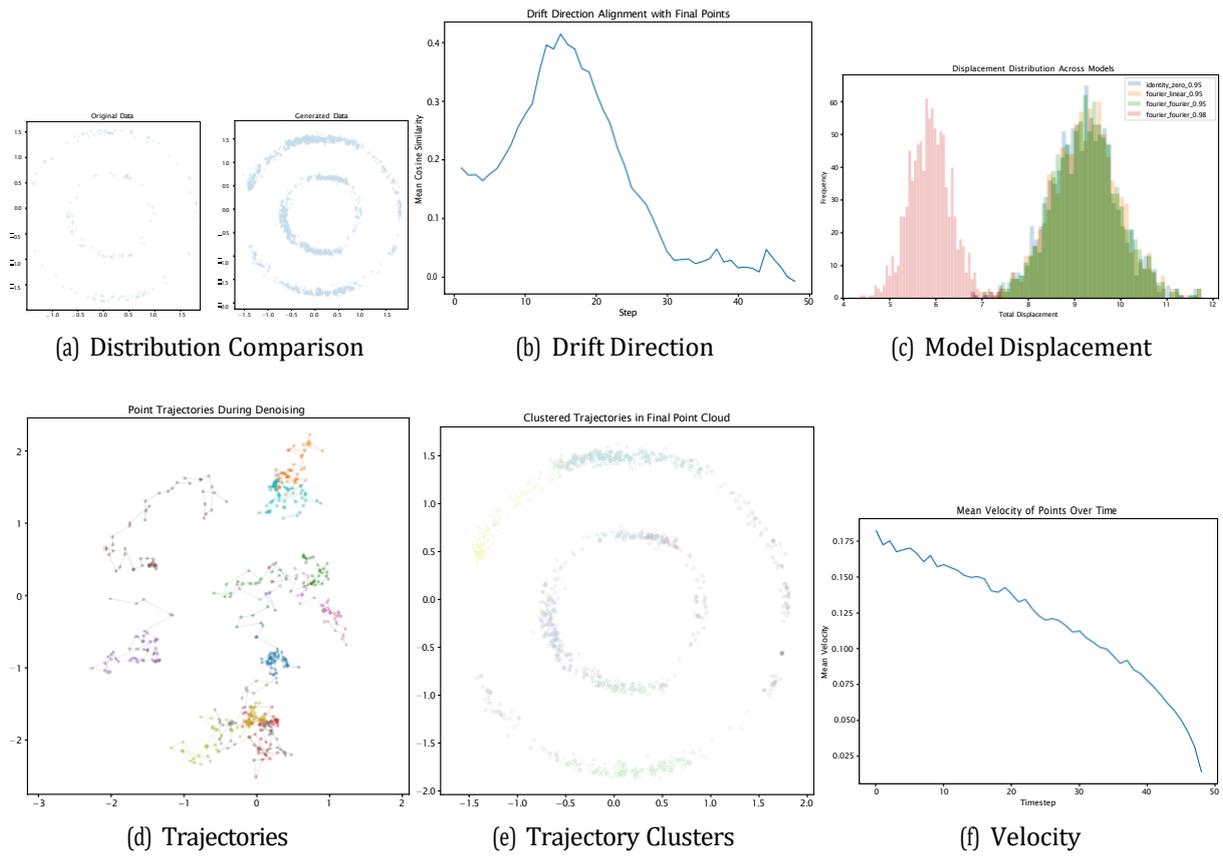

Figure 1: Comparison of different dynamics-related outputs for Bullseye dataset.

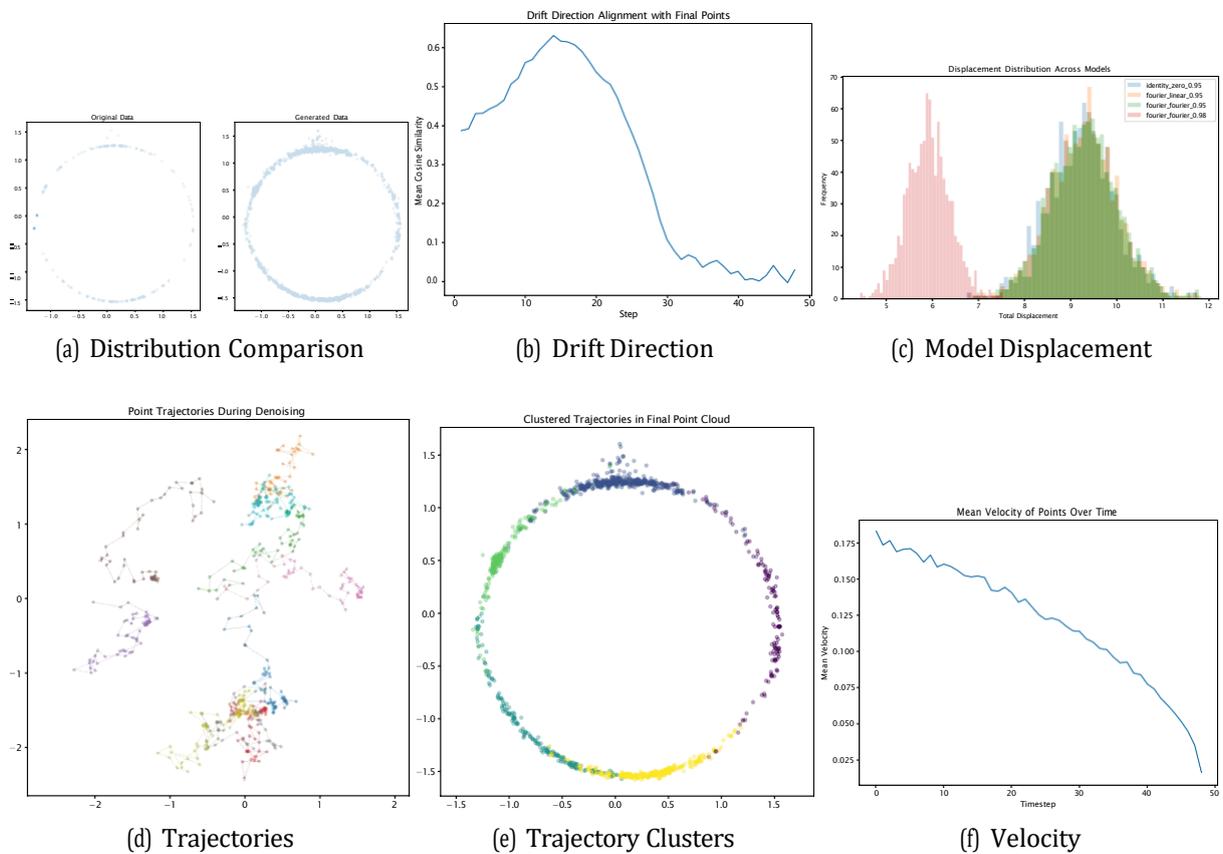

Figure 2: Comparison of different dynamics-related outputs for Circle dataset.

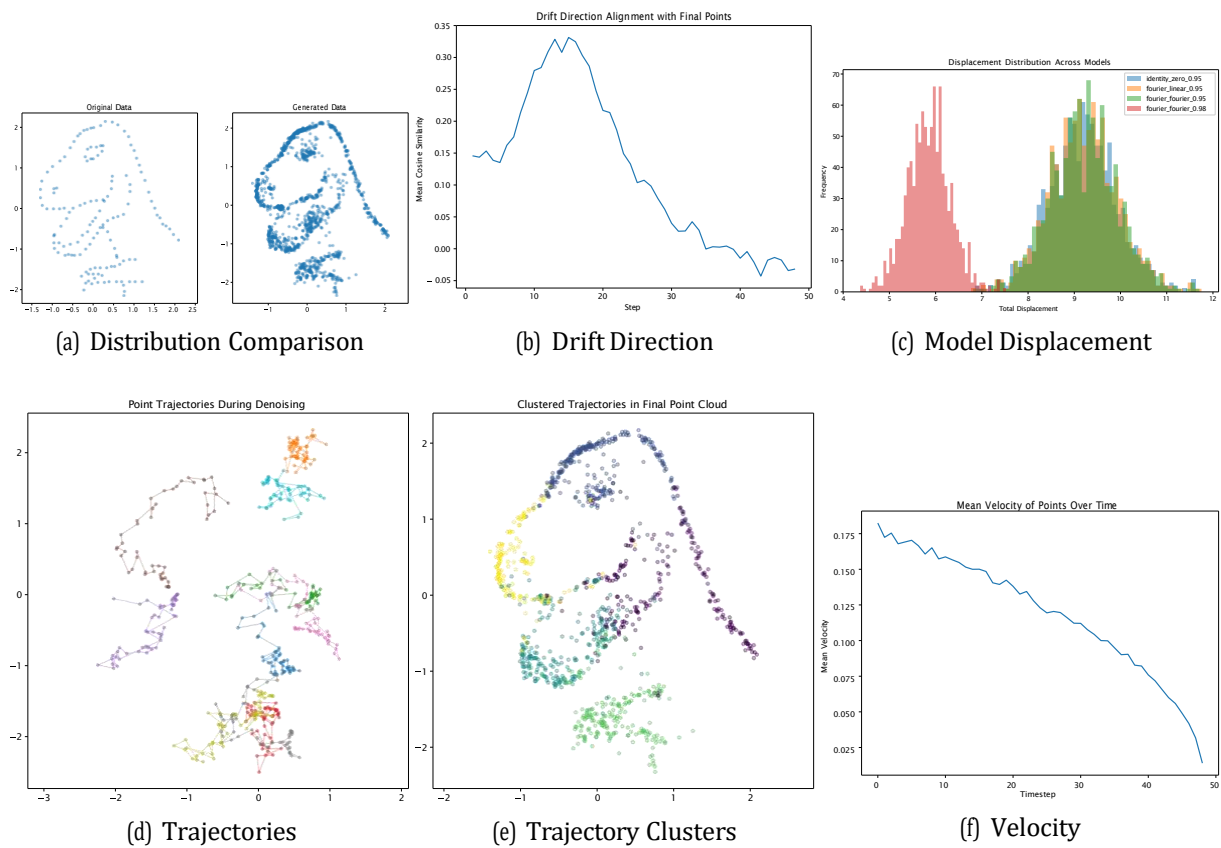

Figure 3: Comparison of different dynamics-related outputs for Dinosaurs dataset.

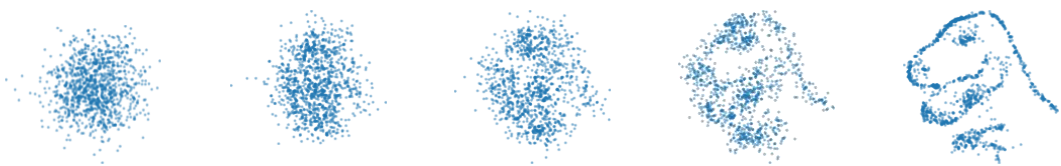

Figure 4: Dinosaurs formation from pure noise with t=10, 20, 30, 40, 50

## A. Drift Direction Explained

The evaluation of drift direction in Figure 1 (b) (for example) is divided into three phases based on the mean cosine similarity between the drift vector and the direction to the final point: in early timesteps with high noise, the similarity is around 0.2, indicating a weak but noticeable alignment and faint pull toward the final state amid chaos; in middle timesteps, it peaks at about 0.4, showing the strongest alignment and effectiveness in guiding samples along the true trajectory as the "sweet spot" of denoising; and in late timesteps with minimal noise, it drops to near 0, becoming nearly orthogonal as the drift shifts to fine-tuning local details rather than directional guidance.